\documentclass[11pt, a4paper, logo, onecolumn,copyright]{tmlrmel}

\usepackage[authoryear, sort&compress, round]{natbib}
\title{Neural Network Reprogrammability: A Unified Theme on Model Reprogramming, Prompt Tuning, and Prompt Instruction}

\usepackage[utf8]{inputenc} 
\usepackage[T1]{fontenc}    
\usepackage{hyperref}       
\usepackage{url}            
\usepackage{booktabs}       
\usepackage{nicefrac}       
\usepackage{microtype}      
\usepackage{amsmath}
\usepackage{graphicx}
\usepackage{multicol}
\usepackage{siunitx}
\usepackage{array}
\usepackage[nameinlink]{cleveref}
\usepackage{bbm}
\usepackage{multirow}
\usepackage{subfig}
\usepackage{soul}
\usepackage{floatrow}
\usepackage{float}
\usepackage{wrapfig}
\usepackage{blindtext}
\usepackage{tablefootnote}
\usepackage{amsfonts}
\usepackage[flushleft]{threeparttable}
\usepackage{bbding}
\usepackage{lipsum}

\theoremstyle{plain}

\theoremstyle{definition}
\newtheorem{definition}{Definition}

\theoremstyle{remark}
\newtheorem{remark}{Remark}
\usepackage{thm-restate}

\usepackage{xcolor}
\usepackage{colortbl}
\definecolor{lg}{gray}{0.9}
\definecolor{dg}{RGB}{0,150,0}
\definecolor{dr}{RGB}{139,0,0}

\usepackage{xspace}

\newfloatcommand{capbtabbox}{table}[][\FBwidth]

\newcommand{\draftonly}[1]{#1}
\newcommand{\eat}[1]{}
\renewcommand{\draftonly}[1]{}
\definecolor{darkgreen}{RGB}{0, 102, 0}

\crefformat{section}{\S#2#1#3}

\usepackage{amsmath,amsfonts,bm}

\def\eqref#1{equation~\ref{#1}}

\def\1{\bm{1}}

\def\rve{{\mathbf{e}}}

\def\rvh{{\mathbf{h}}}

\def\rvx{{\mathbf{x}}}
\def\rvy{{\mathbf{y}}}

\def\mE{{\bm{E}}}

\DeclareMathAlphabet{\mathsfit}{\encodingdefault}{\sfdefault}{m}{sl}
\SetMathAlphabet{\mathsfit}{bold}{\encodingdefault}{\sfdefault}{bx}{n}

\def\gC{{\mathcal{C}}}
\def\gD{{\mathcal{D}}}
\def\gE{{\mathcal{E}}}

\def\gH{{\mathcal{H}}}

\def\gL{{\mathcal{L}}}
\def\gM{{\mathcal{M}}}

\def\gO{{\mathcal{O}}}

\def\gX{{\mathcal{X}}}
\def\gY{{\mathcal{Y}}}

\def\sD{{\mathbb{D}}}

\def\sR{{\mathbb{R}}}

\newcommand{\E}{\mathbb{E}}

\correspondingauthors{fengliu.ml@gmail.com}
\equalcontributions

\author[$\diamondsuit$,*]{Zesheng Ye}
\author[$\diamondsuit$,*]{Chengyi Cai}
\author[$\diamondsuit$,*]{Ruijiang Dong}
\author[$\diamondsuit$]{Jianzhong Qi}
\author[$\spadesuit$]{Lei Feng}
\author[$\clubsuit$]{Pin-Yu Chen}
\author[$\diamondsuit$]{Feng Liu}

\affil[ \hspace{-0.2em}]{$\diamondsuit$ University of Melbourne
}
\affil[ \hspace{-0.2em}]{$\spadesuit$ Southeast University}
\affil[ \hspace{-0.2em}]{$\clubsuit$ IBM Research}

\begin{abstract}
  As large-scale pre-trained foundation models continue to expand in size and capability, efficiently adapting them to specific downstream tasks has become increasingly critical. 
  Despite substantial research progress, existing adaptation approaches have evolved largely in isolation, without a clear understanding of their interrelationships.
  This survey introduces \emph{neural network reprogrammability} as a unifying framework that bridges mainstream model adaptation techniques--model reprogramming, prompt tuning, and prompt instruction--previously fragmented research areas yet converges on a shared principle: repurposing a pre-trained model by \emph{manipulating information at the interfaces} while keeping the model parameters frozen.
  These methods exploit neural networks' sensitivity to information manipulation on different interfaces, be it through perturbing inputs, inserting tokens into intermediate layers, or providing task-specific examples in context, to redirect model behaviors towards desired outcomes.
  Building on the concept of reprogrammability, we present a taxonomy that categorizes such information manipulation-based adaptation approaches across four key dimensions: manipulation format (fixed or learnable), location (interfaces where manipulations occur), operator (how they are applied), and output alignment requirement (post-processing needed to align outputs with downstream tasks). 
  Notably, this framework applies consistently across data modalities, independent of specific model architectures.
  Moreover, viewing established techniques such as {\em in-context learning} and {\em chain-of-thought} prompting through this lens reveals both their theoretical connections and practical distinctions. 
  We further analyze remaining technical challenges and ethical considerations, positioning neural network reprogrammability as a fundamental paradigm for efficient model adaptation.
  We lastly identify promising research directions emerging from this integrative viewpoint.
\end{abstract}

\begin{document}

\maketitle

\section{Introduction}

The rise of foundation models is revolutionizing the landscape of modern machine learning.
By learning general-purpose data representations, these powerful pre-trained architectures have delivered unprecedented results in computer vision~\citep{dosovitskiy2021image, zhao2023survey} and natural language processing~\citep{vaswani2017attention, devlin2019bert, brown2020language}, but their massive scale--often billions of parameters--creates significant deployment challenges.

\begin{figure}[t]
    \centering
    \includegraphics[width=\linewidth]{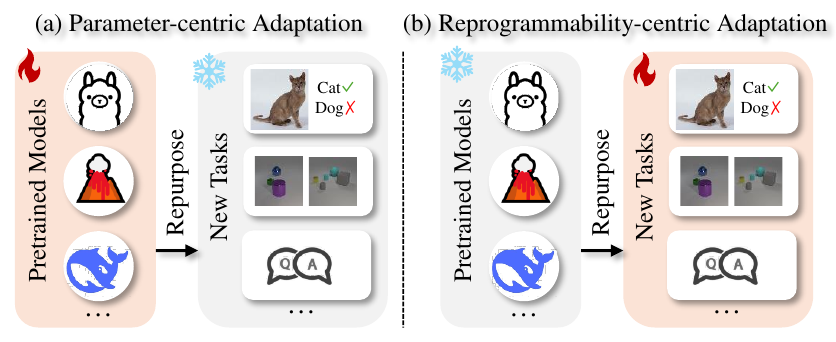}
    \caption{Paradigm shift from conventional \emph{\textcolor{brown}{parameter-centric adaptation}} (i.e., modifying model parameters) to \emph{\textcolor{purple}{reprogrammability-centric adaptation}} (i.e., modifying input data and model output). 
    This represents a shift in thought from {\em {\color{brown} modifying the model to align with the task}} to {\em {\color{purple} modifying the task to align with the model}}. 
    }
    \label{fig:intro}
\end{figure}

Conventionally, adapting these models to specific downstream tasks necessitates an intensive process of \emph{fine-tuning}, where a substantial portion, if not all, of the model's \emph{parameters} are updated using task-specific data~\citep{pan2009survey, chen2024model}. 
This \textcolor{brown}{{\em parameter-centric adaptation strategy}}~(PCA), depicted in Figure~\ref{fig:intro} (a), directly modifies the model's learned weights.
While straightforward and often effective, PCA may come with several limitations, especially when resources are constrained.
The computational demands of retraining these models often become prohibitively expensive. 
Access to model internals may not be available often when models are provided as services (i.e., black-box setting). 
Furthermore, the need for specialized hardware (e.g., high-performance GPUs) creates barriers to entry, and the significant energy footprint raises valid environmental concerns.

A more efficient paradigm is gaining traction.
Rather than {\color{brown} modifying the internal parameters}, a growing body of work now focuses on adapting pre-trained models for downstream tasks (i.e., target tasks) {\color{purple} without} extensive (or, in some cases, any) {\color{purple} parameter updates}. 
As Figure~\ref{fig:intro} (b) shows, these approaches strategically manipulate the target task-specific inputs or contextual information provided to a {\color{blue} fixed} pre-trained model, and then align the model output to the requirement of the target task's output space.
In this paper, we term such methods \emph{\textcolor{purple}{reprogrammability-centric adaptation}} (RCA) and formally define the concept of {\em reprogrammability} in Sec.~\ref{sec: adversarial_vulnerability}.
Importantly, the shift of adaptation focus in RCA methods leads to a compelling benefit: model adaptation costs (in terms of trainable parameters) become largely independent of model size, correlating instead with the complexity of the task itself.
Namely, the input and output dimensionality of the target data.

\begin{figure}[t]
    \centering
    \includegraphics[width=0.95\linewidth]{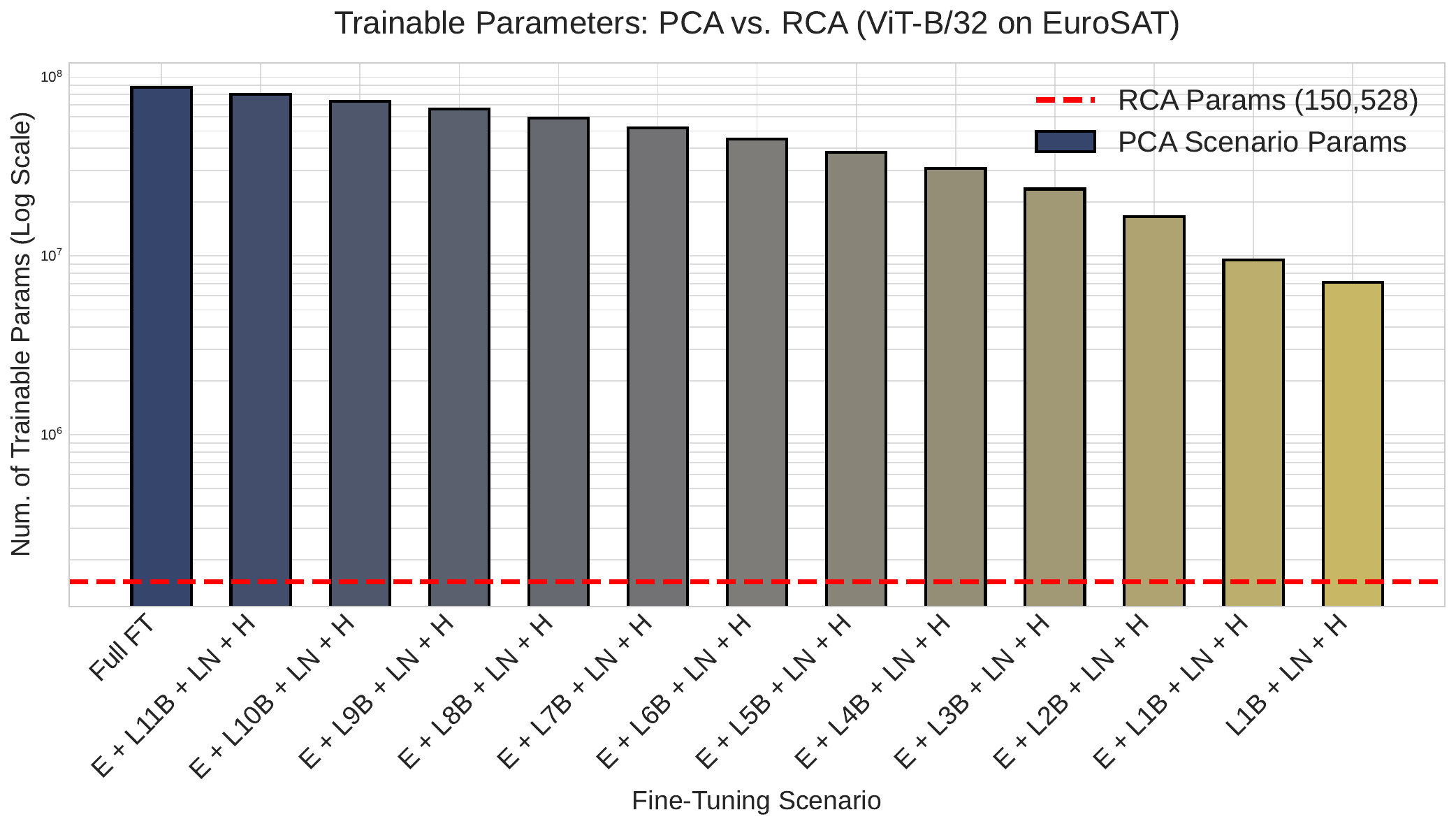}
    \caption{Comparison of trainable parameters for fine-tuning (i.e., PCA) vs. reprogramming (i.e., RCA) an ImageNet-pretrained ViT/B-32 on a new classification task. Different PCA scenarios are shown (e.g., "E+L10B+LN+H" indicates fine-tuning the embedding layer, the last 10 Transformer blocks, layer normalization, and the classification head). The dashed line indicates the number of trainable parameters for RCA, which is consistently lower than that of most PCA configurations.}
    \label{fig:efficiency_comparison}
\end{figure}

The efficiency benefits can be substantial.
Consider a typical neural network, such as a {\em multi-layer perceptron}~(MLP) with $L$ layers of dimension $D$ processing inputs of dimension $d$, we typically find that $D \times L \gg d$--the model's complexity substantially exceeds that of the data it processes since neural networks are heavily over-parameterized~\citep{zhang2016understanding, zhu2019learning}.
This implies that RCA methods can reduce the training burden as they shift the model adaptation efforts from the parameter space to a much lower-dimensional input space.
Figure~\ref{fig:efficiency_comparison} showcases a concrete illustration, where we compare the number of trainable parameters when adapting an ImageNet-pretrained ViT/B-32~\citep{dosovitskiy2021image} to a new remote sensing image classification task \citep{eurosat}.
The comparison is between various fine-tuning (i.e., {\color{brown} PCA}) scenarios and reprogramming (i.e., {\color{purple} RCA}), confirming that RCA consistently involves significantly fewer trainable parameters than most PCA scenarios.
The implications of this efficiency are far-reaching.
By significantly reducing computational demands, RCA methods promise to democratize access to cutting-edge AI capabilities, enable deployment in resource-constrained environments, and reduce the carbon footprint of AI applications.

\subsection{Motivation: The Need for Terminological Clarity}
Despite the promise of this emerging paradigm, a significant obstacle to progress in this field is the bewildering array of inconsistent terminology.
Techniques like ``prompt tuning'' and ``model reprogramming'' and variations thereof are often used interchangeably or, worse, with conflicting definitions across research communities, such as the same terms applied to fundamentally different approaches.
Consider how the term ``prompt tuning'' operates in the literature.
In some contexts, it broadly encompasses any modifications to the raw input data~\citep{brown2020language}, yet in other contexts, it specifically refers to adding trainable continuous vectors to input embeddings~\citep{li2021prefix, lester2021power}.
Similarly, ``model reprogramming'' can denote input transformations~\citep{elsayed2019adversarial, chen2024model} and, elsewhere, describe manipulations within embedding spaces~\citep{neekhara2019adversarial}.
This terminological ambiguity creates unnecessary barriers to knowledge transfer across research communities. 
Accordingly, researchers who work on similar problems may struggle to identify and build upon each other's work or conduct fair comparisons, ultimately slowing progress in such a field with vast potential to broaden access to state-of-the-art machine learning capabilities.

\subsection{Purpose: Unifying Disparate Approaches}
To address these issues, this survey introduces \emph{Neural Network Reprogrammability} as a \emph{unifying conceptual framework} to bring clarity and structure to the diverse landscape of parameter-efficient adaptation methods.
We define {\em reprogrammability} as the inherent capability of pre-trained models to be repurposed for new tasks {\em without internal parameter updates}; whilst the adaptation is achieved by strategically manipulating the inputs (and/or auxiliary) information provided to the model~\citep{elsayed2019adversarial, chen2024model} and the output model produces, effectively leveraging neural networks' inherent input-output sensitivity~\citep{lowd2005adversarial} to elicit desired task-specific behaviors.

\begin{figure}[t]
    \centering
    \includegraphics[width=0.8\linewidth]{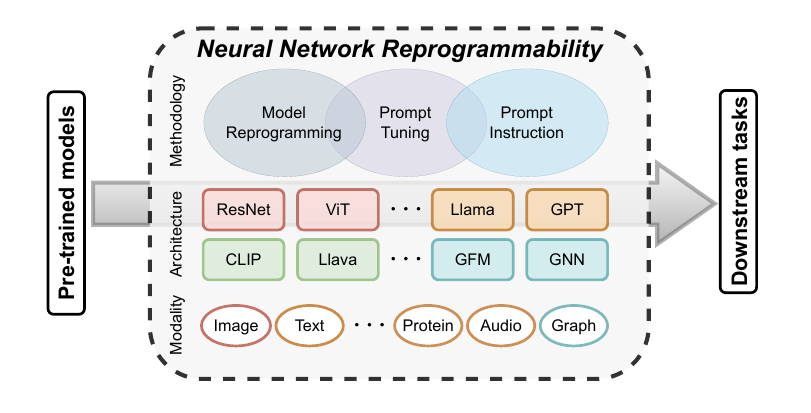}
    \caption{We introduce \emph{Neural Network Reprogrammability} as a unifying framework to bring coherence to a set of model adaptation techniques that have often been studied in isolation. Importantly, this shared underlying principle applies regardless of pre-trained model architectures and data modalities.}
    \label{fig:unified}
\end{figure}

Reprogrammability provides a coherent lens through which to view three major families of work: \emph{model reprogramming} (MR)~\citep{elsayed2019adversarial, chen2024model}, \emph{prompt tuning} (PT)~\citep{lester2021power, li2021prefix}, and \emph{in-context learning}~\citep{brown2020language, kirillov2023segment}, which we refer to as \emph{prompt instruction}~(PI) throughout this paper to avoid confusion with PT.
Although often developed in parallel across different communities, we argue that they all rely on a common underlying principle that has yet to be formally recognized and consolidated. 
This commonality motivates grouping them under the unifying concept of \emph{\textcolor{purple}{reprogrammability-centric adaptation}}~(RCA). 
This perspective reveals the connections between seemingly disparate methods, showing them as variations on a common theme: the systematic exploitation of a neural network's sensitivity to manipulations at its interfaces.

With this unified view (Figure~\ref{fig:unified}), we aim to bridge previously disparate research efforts by uncovering the overarching principles of adaptation that hold across different techniques, data modalities, and applications. 
By examining these methods through the common lens of {\em neural network reprogrammability}, we intend to clarify their relationships, foster more precise cross-community terminology and discussions, and stimulate deeper investigations into the core mechanisms. 
Ultimately, this should accelerate the creation of more effective, efficient, and broadly accessible means for adapting large-scale pre-trained models.

\subsection{Scope and Contributions of this survey}

Before outlining our contributions, we first acknowledge the extensive body of existing work and delineate the scope of this survey with existing literature.
The field has seen remarkable survey papers covering specific methodologies and applications in detail, from model reprogramming~\citep{chen2024model} and comprehensive reports on the prompting technique~\citep{schulhoff2024prompt}, to surveys on textual~\citep{luo2024incontext} and visual~\citep{zhang2023survey} in-context learning, and prompt engineering for visual-language models~\citep{gu2023systematic} or even broader multi-modal architectures~\citep{wu2024visual}.
To avoid redundancy and maintain a clear focus, our work will not replicate these specialized reviews.

\begin{table}[t]
\caption{
Comparison with existing surveys related to reprogramming and prompting techniques.
Unlike existing surveys that focus on specific modalities, applications, model architectures, or benchmarks, this work aims to systematically organize the diverse (and sometimes miscellaneous) methodologies and terminologies under a unified conceptual framework with a clear taxonomy. 
}
\resizebox{\linewidth}{!}{%
\begin{tabular}{llllllll}
\toprule
\textbf{Survey Focus} & \citet{gu2023systematic} & \citet{schulhoff2024prompt} & \citet{luo2024incontext} & \citet{zhang2023survey} & \citet{wu2024visual} & \citet{chen2024model} & Ours \\
\midrule
\textbf{In-Modal Progress}                 & \checkmark     & \checkmark        & \checkmark    & \checkmark & \checkmark    &         &      \\
\textbf{Application-targeted}              & \checkmark     & \checkmark        & \checkmark    & \checkmark & \checkmark    &         &      \\
\textbf{Arch-agnostic}          & \checkmark     &            &        &     & \checkmark    & \checkmark     & \checkmark  \\
\textbf{Cross-Field}           &         & \checkmark        &        &     &        & \checkmark        & \checkmark  \\
\textbf{Unifying Paradigm}        &         &            &        &     &        &         & \checkmark \\
\bottomrule
\end{tabular}
}
\label{tab:comparison_survey}
\end{table}

\paragraph{Relation to Existing Surveys.}
While several recent surveys have covered related areas, such as prompting or parameter-efficient fine-tuning~\citep{gu2023systematic, schulhoff2024prompt, zhang2023survey, chen2024model}, this survey distinguishes itself by offering a distinct perspective and presenting \emph{neural network reprogrammability} as a central concept to unite different adaptation families (MR, PT, PI) that have no internal parameter updates.
Existing surveys typically focus on tracking advancements within a specific technique (e.g., prompting \textbf{or} model reprogramming) \textbf{or} application area. 
In contrast, we propose a more general framework (Section~\ref{sec: framework}) that spans traditionally separate research communities, along with a systematic taxonomy (Section~\ref{sec: taxonomy}) that classifies RCA methods along four key dimensions.
This helps reveal connections that remain potentially hidden when these approaches are studied in isolation (See Table~\ref{tab:comparison_survey} for a comparative summary).

\paragraph{Contributions.}
This paper presents a structured overview for adapting pre-trained models without updating their parameters. 
We introduce \emph{neural network reprogrammability} as the core concept and a \textbf{new} perspective to unify what has become a fragmented and often terminologically confusing area of research.
We connect parameter-efficient adaptation techniques used across diverse modalities, emphasizing their shared strategy of input/output manipulation rather than parameter modification.
This survey makes four primary contributions:

\begin{itemize}
    \item[1)] \textbf{A Conceptual Unification.} 
    We show that model reprogramming, prompt tuning, and prompt instruction--techniques often developed independently--share fundamental principles.
    They all exploit pre-trained representations and guide model behaviors by carefully crafting inputs and outputs.
    We provide a conceptual framework that bridges these distinct lines of study.
    
    \item[2)] \textbf{A Modality-Agnostic Taxonomy.}
    We develop a systematic way to classify reprogramming and prompting techniques that transcends specific domains (text, images, etc.) or architectures, such as ResNet~\citep{he2016deep}, ViT~\citep{dosovitskiy2021image}, BERT~\citep{devlin2019bert}, and CLIP~\citep{radford2021learning}. 
    Namely, this taxonomy organizes RCA methods based on \emph{what} (i.e., whether manipulations are ``hard'' or ``soft'' ones), \emph{how} (i.e., which operator is relied on to insert manipulations), and \emph{where} (i.e., locations where manipulations are applied to) the input or context is manipulated, as well as \emph{whether} post-processing is required to align pre-trained model outputs with target tasks. 
    This taxonomy enables a clear comparison of different RCA approaches.
    
    \item[3)] \textbf{A Comparative Analysis.}
    We place the proposed unified framework RCA within a broad context of model adaptation research, highlighting connections and differences with other approaches.
    Unlike surveys focused narrowly on specific techniques or domains, this paper examines how reprogrammability enables model reuse across diverse tasks and architectures and touches upon underlying theoretical aspects typically overlooked by previous studies.

    \item[4)] \textbf{A Discussion of Open Challenges and Future Directions.}
    We identify fundamental open questions regarding the theoretical foundations of reprogrammability, analyze limitations of current evaluation methods, and discuss emerging ethical issues, aiming to catalyze future research in this rapidly evolving field.
\end{itemize}

\section{A Unified View of Reprogrammability}
\label{sec: framework}

Here, we define reprogrammability as the capacity of  \textcolor{cyan}{\emph{fixed} neural network} to be redirected toward tasks outside its original training task, by {\color{red} {\em manipulations}} within its {\color{red} input and output spaces}, such that the new task's context can fit in the pre-trained model's learned feature space.
This concept unifies a range of parameter-efficient adaptation techniques by highlighting their shared reliance on the model's intrinsic sensitivity to input and/or output transformations.
Later, in Sec.~\ref{sec: taxonomy}, we will show how existing model adaptation strategies fit this definition and can be categorized along four dimensions: (1) the {\em format} of the manipulation, (2) its {\em location} within the model, (3) the {\em operator} used, and (4) the need for {\em output alignment}.

\subsection{From Vulnerability to Adaptation}
\label{sec: adversarial_vulnerability}

\paragraph{Input Sensitivity.}
The remarkable capabilities in end-to-end representation learning of neural networks are juxtaposed by a paradoxical property: their susceptibility to adversarial manipulation~\citep{goodfellow2014explaining, carlini2017adversarial, chen2018ead}.
Studies in adversarial machine learning have consistently shown that even high-performing models can be misled by carefully crafted input perturbations that remain imperceptible to humans~\citep{lowd2005adversarial, carlini2017towards}. 
This vulnerability is not merely an artifact of specific architectures or training procedures but appears to be rooted in the fundamentally discontinuous way neural networks map high-dimensional input spaces to outputs~\citep{szegedy2013intriguing}.

Several theories have attempted to explain this phenomenon.
One hypothesis attributes it to the locally linear behavior of neural components, where small input changes can accumulate across layers, leading to large shifts in the final output~\citep{goodfellow2014explaining}.
Another perspective points to the geometry of high-dimensional spaces, where decision boundaries, while potentially far from data points in low dimensions, can become unexpectedly close in high dimensions, making them fragile to even subtle perturbations in the input space~\citep{fawzi2016robustness, gilmer2018adversarial}.
Additionally, recent work suggests models might rely heavily on ``non-robust features''--patterns statistically correlated with labels but easily altered adversarially--which contribute significantly to standard generalization but also to adversarial fragility~\citep{ilyas2019adversarial}.
Such inherent input sensitivity means the learned function, while accurate on the training data distribution, possesses complex and sensitive decision boundaries when exploring regions slightly off the training data manifold \citep{elsayed2019adversarial, chen2024model}.

\noindent
\textbf{Repurposing Sensitivity for Adaptation.}
This inherent input sensitivity, conventionally framed as a weakness, can however be reinterpreted as the \emph{enabler of neural network reprogrammability}. 
The very mechanisms that allow adversarial examples to drastically alter model predictions--namely, the ability to find specific input directions that maximally change the output--can be repurposed. 
Thus, a seemingly detrimental vulnerability can be transformed beneficially. 
Instead of perturbing inputs to make a model fail (e.g., via gradient ascent on the loss function), we can engineer input transformations to guide a pre-trained model to perform tasks it was not originally designed or trained for.
Such transformations eventually form the foundation of neural network reprogrammability.

Specifically, reprogrammability allows us to reuse a model's established computational graph and learned feature hierarchy.
Consider an image classifier pre-trained on ImageNet~\citep{deng2009imagenet}, which has learned a set of visual features including edges, textures, parts, and objects.
Reprogrammability seeks to leverage this existing knowledge by learning transformation function(s) at the model's interfaces--input, intermediate representations, and/or output--that map data from the new task (say, classifying medical images) into a format that effectively makes use of these pre-learned features and (potentially) then map model predictions to align with new label spaces.
For instance, one might learn an input transformation that adds a particular pattern to medical images, causing the classifier to activate representations useful for the new task, combined with a mapping from the ImageNet classes to the desired medical labels~\citep{chen2024model}.

This viewpoint marks a paradigm shift in model adaptation. 
While traditional transfer learning methods, e.g., fine-tuning~\citep{pan2009survey, guo2022deep} and domain adaptation~\citep{long2016unsupervised, long2018conditional}, which focus on modifying weights of the source model, have been the de-facto practices, these approaches, especially the latter ones, find challenging source-free domain adaptation setting~\citep{liang2020we, chi2021tohan}, where adaptation to a new target domain must be performed using only the pre-trained source model and target domain data, without any access to the original source data\footnote{Many source-free methods still rely on updating parts of the model, such as batch normalization statistics or select layers, often through techniques like pseudo-labeling or entropy minimization~\citep{dong2023diversity}.
This places them in a distinct category from the reprogrammability-based methods we discussed here.}.
By contrast, with reprogrammability, models can be instead adapted by manipulating the data fed to a \emph{\textcolor{blue}{fixed}} function.
This implies that the emphasis shifts from updating the model to transforming the data.
Through careful design of transformations, we can effectively ``reprogram'' a model's input-output mapping, without touching its core learned representations.

\subsection{Formulation of Reprogrammability}

We now formalize neural network reprogrammability to provide a precise framework for understanding diverse RCA approaches. 
This formulation captures how a fixed model can be repurposed for new tasks through strategic transformations at its interfaces.

\begin{definition}[Neural Network Reprogrammability]\label{def: reprogrammability}
    Let $f(\rvx^{\rm S}; \theta)$ be a pre-trained neural network with parameters $\theta$, representing a learned mapping $f: \gX^{\rm S} \to \gY^{\rm S}$ from the input space $\gX^{\rm S}$ to the output space $\gY^{\rm S}$ in a source (i.e., pre-trained) domain $\gD^{\rm S} \subseteq \gX^{\rm S} \times \gY^{\rm S}$.
    We consider \emph{neural network reprogrammability} as the realization of $f(\rvx; \theta)$ that induces a \emph{target functionality}, jointly defined over the target input and output spaces $\gX^{\rm T} \times \gY^{\rm T}$.
    The reprogramming process comprises two \emph{configurable mappings}.
    Namely, an {\em input manipulation} mapping $I_{\lambda, \tau, \ell}$ applied prior to $f$, and an {\em output alignment} mapping $O_{\omega}$ applied after $f$.
\end{definition}

\paragraph{Input Manipulation.} 
The central mechanism for adapting the pre-trained model $f$ is {\em input manipulation} $I_{\lambda, \tau, \ell}: \gX^{\rm T} \times \gC \to \gM$, which transforms target input $\gX^{\rm T} \in \gX^{\rm T}$ before it, or its subsequent representations, are processed by parts of $f$.
The manipulation occurs at one or multiple of the model's interfaces $\gM = \{ \gX^{\rm S}, \gE, \gH \}$, which can be its raw input space $\gX^{\rm S}$, its embedding space $\gE$, or its deeper hidden representation spaces $\gH$\footnote{When manipulations occur at embedding or hidden layers, it is helpful to view these layers as processing units whose own inputs are being altered.}.
Moreover, the manipulation can potentially be guided by contextual information $\gC$, which refers to task-specific contexts and constraints guiding how the pre-trained model should be repurposed.
This transformation is characterized by
    \begin{itemize}
        \item \textbf{Manipulation Configuration} $\lambda \in \Lambda$ that specifies the transformation parameters.
        We partition $\Lambda = \Lambda_{\rm fixed} \cup \Lambda_{\rm learnable}$, distinguishing between fixed configurations (i.e., ``hard prompts'') and learnable parameters (i.e., ``soft prompts'').
        For learnable configurations, $\lambda$ is typically optimized by minimizing a task-specific objective function $\gL: \gY^{\rm T} \times \gY^{\rm T} \to \sR^{+}$.
        \item \textbf{Manipulation Location} $\ell \in \gM$ that specifies where in the computational graph the manipulation occurs, including three options: Input layer $\ell_0$, i.e., modifications applied directly to raw input space $\gX^{\rm S}$; embedding layer $\ell_1$, i.e., modifications applied to the embedding space $\gE$; intermediate layers $\ell_i$ for $i = \{ 2, \dots, n \}$, i.e., modifications applied to the hidden representation space $\gH$.
        \item \textbf{Manipulation Operator} 
        $\tau: \gX^{\rm T} \times \lambda \to \gM$ that defines how manipulation is incorporated along with the target task data, and $\tau \in \{{\rm add}, {\rm concat}, {\rm param} \}$.
        Generally, the operator can be addition $\tau_{\rm add} (\rvx^{\rm T}, \lambda) = \rvx^{\rm T} + \lambda$ in cases where $\rvx^{\rm T}$ and $\lambda$ are appropriately dimensioned, or concatenation $\tau_{\rm concat}(\rvx^{\rm T}, \lambda) = \left[\rvx^{\rm T}; \lambda \right]$, denoting the concatenation along a specified dimension, or parametric transformation $\tau_{\rm param}(\rvx^{\rm T}, \lambda) = \lambda(\rvx^{\rm T})$ with a function $\lambda(\cdot)$ that maps target input to a space compatible with the input space of the corresponding interface.
    \end{itemize}
    Moreover, multiple manipulations can be composed sequentially as $I_{\lambda, \tau, \ell} \triangleq I_{\lambda_n, \tau_n, \ell_n} \circ \cdots \circ I_{\lambda_1, \tau_1, \ell_1}$, enabling sophisticated manipulation strategies that leverage different model interfaces.
    For this composition to be well-defined, the output space of $I_{\lambda_i, \tau_i, \ell_i}$ needs to be compatible with the input space of $I_{\lambda_{i+1}, \tau_{i+1}, \ell_{i+1}}$ for all $i \in \{1, \dots, n-1 \}$ throughout the computational graph.
    Formally, $I_{\lambda, \tau, \ell}$ induces a mapping that transforms target domain input $\rvx^{\rm T} \in \gX^{\rm T}$ and context $c \in \gC$ to a sub-manifold $\gM^{\prime} \subset \gM$ that intersects meaningfully with the effective feature spaces learned by $f$ during its pre-training.
        
\paragraph{Output Alignment.} 
Once the pre-trained model $f$ processes the manipulated input to produce a source-domain output $y^{\rm S} \in \gY^{\rm S}$, this output often needs to be mapped to the format compatible with the target task.
The output alignment mapping $O_\omega: \gY^{\rm S} \to \gY^{\rm T}$, parametrized by configuration $\omega \in \Omega$, this function performs this step, establishing necessary correspondences between source and target output spaces $\gY^{\rm S}$ and $\gY^{\rm T}$.
Key alignment methods include:
\begin{itemize}
    
    \item \textbf{Identity Mapping}. 
    The simplest approach, used when $\gY^{\rm S}$ is directly compatible with $\gY^{\rm T}$, i.e., $\gY^{\rm S} = \gY^{\rm T}$, requiring no explicit alignment transformation. This is common in generative tasks where the model directly produces outputs in the desired format.
    
    \item \textbf{Structured Alignment}. 
    This involves a mapping $O_\omega ( \rvy^{\rm S}) = \omega(\rvy^{\rm S})$, where $\omega(\cdot)$ is typically a pre-defined, fixed function (e.g., a rule-based parser or extractor), aiming to isolate or reformat task-relevant information from the raw output of the pre-trained model, often employed for tasks requiring specific output structures.
    
    \item \textbf{Statistical Alignment}. 
    This involves a mapping $O_\omega ( \rvy^{\rm S}) = \omega \rvy^{\rm S}$, where $\omega \in \{0, 1\}^{|\gY^{\rm T}| \times |\gY^{\rm S}|}$ is a binary matrix. The matrix $\omega$ is determined non-parametrically based on statistical correspondences between the model's source predictions and the target labels, known as {\em label mapping}~\citep{tsai2020transfer, chen2023understanding}.
    
    \item \textbf{Linear Alignment}. 
    This also uses a matrix transformation $O_\omega ( \rvy^{\rm S}) = \omega \rvy^{\rm S}$, but here $\omega \in \sR^{|\gY^{\rm T}| \times |\gY^{\rm S}|}$ is a real-valued matrix. 
    The matrix $\omega$ can be obtained either by gradient-based backpropagation~\citep{tsao2023autovp}, e.g., optimizing a linear probe trained on top of the frozen pre-trained model's outputs; or statistically estimated without backpropagation, for example, by estimating the optimal probabilistic mapping matrix from observed statistics of source predictions and target labels \citep{cai2024bayesian}\footnote{We note that \citet{cai2024bayesian} also relies on statistics and does not rely on gradient-based optimization. 
    Still, here we distinguish it from the ``statistical alignment'' category as $\omega$ is not restricted to binary values and can represent more precise linear transformations.}.
\end{itemize}

The choice of output alignment method depends heavily on the nature of the target task and the relationship between the source and target output spaces.
linear projections and class remappings are often set for downstream classification tasks, whilst structured extraction is utilized for complex tasks like language modeling.

Without loss of generality, the \emph{reprogrammed neural network} can be represented as 
\begin{equation}\label{eq: reprogrammed_neural_net}
    f^{\prime}(\rvx^{\rm T}; \theta, \lambda, c, \ell, \tau, \omega) \triangleq O_\omega \circ f \circ I_{\lambda, \tau, \ell} \, \left(\rvx^{\rm T}, c \right),
\end{equation}
by bringing together both transformation components.

Importantly, this composition achieves the target functionality \emph{without modifying} $\theta$.
In practice, the adaptation process focuses entirely on determining optimal values for contextual guidance $c \in \gC$, manipulation configurations $\lambda \in \Lambda$, and output alignment parameters $\omega$.
These configuration elements may be manually specified using domain expertise or learned through optimization against task-specific objectives.
The implementation varies across adaptation method families, which we will detail in Sec.~\ref{sec: taxonomy}.

\begin{figure*}[t]
    \centering
    \includegraphics[width=\linewidth]{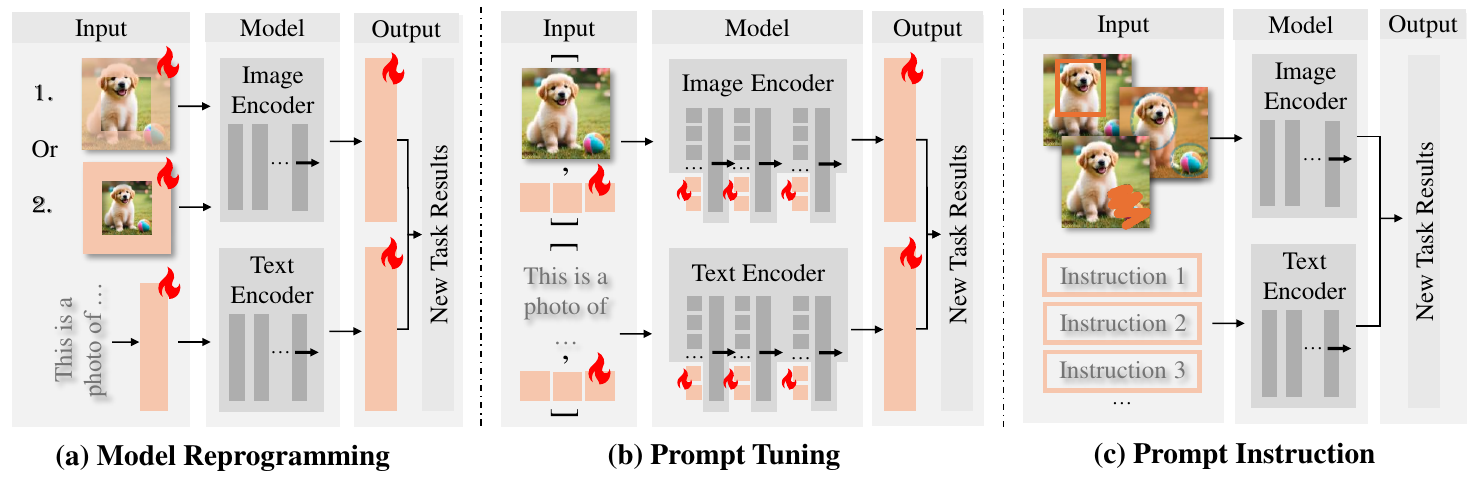}
    \caption{Comparisons of implementations between MR, PT, and PI (using a pre-trained vision-language model as the example), distinguishing from each other by different manipulation forms, locations, operators, and output alignment strategies.}
    \label{fig:methods}
\end{figure*}

\subsection{Revisit Existing Manifestations of Reprogrammability}
The proposed framework (Definition~\ref{def: reprogrammability}) provides a lens through which we can understand several popular efficient adaptation families--\textit{model reprogramming} (MR), \textit{prompt tuning} (PT), and \textit{prompt instruction} (PI)-- as variations on a central theme.
Despite emerging from different research communities and contexts, they primarily diverse in how they implement input manipulation $I(\cdot; \lambda, \tau, \ell, c)$ and output alignment $O(\cdot; \omega)$.
Figure~\ref{fig:methods} visualizes these differences, and Table~\ref{tab: framework_manifestation} summarizes a comparison of these approaches.

\paragraph{Model Reprogramming~(MR).} 
This represents an early structured approach~\citep{chen2024model, elsayed2019adversarial} to repurpose pre-trained models without changing their weights, redirects model processing pathways through task-specific input perturbations to achieve desired outputs on a new target task.
MR instantiates Eq.~(\ref{eq: reprogrammed_neural_net}) by applying manipulation at either the input (for continuous data) or embedding interfaces (for discrete data), formally $\ell= \{ \gX^{\rm S}, \gE \}$, with optimizing $\lambda \in \Lambda_{\rm learnable}$ against task-specific objective functions.

For continuous inputs such as images, MR commonly equips three kinds of manipulations, namely (1) \textit{additive perturbation} $I_{\lambda, {\rm add}, \gX^{\rm S}} = {\rm reshape}(\rvx^{\rm T}) + \lambda$ that superimposes learnable patterns onto reshaped inputs, preserving dimensionality while modifying feature activations; (2) \textit{concatenative augmentation} $I_{\lambda, {\rm concat}, \gX^{\rm S}} = \left[\rvx^{\rm T}; \lambda_{\rm pad}\right]$, extending the input dimensionality by concatenating learnable elements alongside the original data; (3) \textit{transformative projection} $I_{\lambda, {\rm transform}, \gX^{\rm S}} = \lambda \left( \rvx^{\rm T} \right)$, a parameterized function $\lambda$ that projects target inputs into a representation aligned with the pre-trained model's input distribution.
For discrete inputs such as text, MR typically operates at the embedding level by integrating learnable token representations into the embedding space. 
These learnable parameters $\lambda$ serve as domain-bridging elements that activate relevant features within the model's representation space, influenced by $c \in \gC$ that encodes contextual information on perturbations.

Since MR usually adapts $f$ to a new task with a different output space, the output alignment $O_{\omega}: \gY^{\rm S} \to \gY^{\rm T}$ is critical. 
In the case of classification tasks for example, $O_{\omega}$ is often implemented as a linear mapping $\omega \in \sR^{|\gY^{\rm T}| \times |\gY^{\rm S}|}$ or a remapping permutation matrix $\omega \in \{0, 1\}^{|\gY^{\rm S}| \times |\gY^{\rm T}|}$, translating the model's output categories to target task categories and making pre-trained model's predictions meaningful in new tasks.

\paragraph{Prompt Tuning (PT).}
Initially developed for language models, now been extended to multimodal architectures, particularly vision-language models, PT~\citep{li2021prefix, lester2021power} also heavily employs learnable manipulations $\lambda$, but differ from MR in the manipulation interfaces.
Often, PT involves prepending learnable tokens to the model's embedding $\rve \in \gE$ or hidden representations $\rvh \in \gH$.

PT instantiates Eq.~(\ref{eq: reprogrammed_neural_net}) with $\tau = {\rm concat}$ and $\ell \in \gE \cup \gH$.
A canonical formulation of input manipulation is $I_{\lambda, {\rm concat}, \gE} = \left[\lambda_c; \mE(\rvx^{\rm T})\right]$, where $\mE(\rvx^{\rm T})$ is the embedding of target input $\rvx^{\rm T}$ and $\lambda_c$ are task-specific prompt embeddings conditioned on the adaptation directive $c$, prepended to the input embeddings in the embedding space.
PT can extend beyond a single interface, manipulating representations across multiple layers throughout the model architecture, such that $I_{\lambda, \tau, \ell}(\rvx^{\rm T}, c) = I_{\lambda_n, {\rm concat}, \gH_n} \circ \cdots \circ I_{\lambda_1, {\rm concat}, \gE}$.
This cascading manipulation allows more fine-grained control over the model's internal processing path.
Output alignment $O_{\omega}$ varies with task specifications, from selective token extraction for generation tasks~\citep{wang2023tuning}, to parameterized classification heads for discriminative tasks~\citep{li2021prefix}.
When source and target output spaces are naturally aligned $\gY^{\rm S} = \gY^{\rm T}$ (e.g., in text generation with language models), $O_{\omega}$ defaults to an identity mapping, requiring no additional transformation.

\paragraph{Prompt Instruction (PI).}
Also known as in-context learning, PI~\citep{brown2020language, kirillov2023segment} guides model behavior through carefully designed textual or visual instructions, without explicit parameter updates (i.e., $\lambda$ is non-trainable).
PI demonstrates that sufficiently large pre-trained models, e.g., large-language models, can perform novel tasks when provided with appropriate instructional contexts, such as task descriptions or input-output examples.
The term ``in-context learning'' arises from the model's ability to learn from examples provided directly within the input context window.

PI instantiates Eq.~(\ref{eq: reprogrammed_neural_net}) with non-trainable configuration $\lambda \in \Lambda_{\rm fixed}$ by employing $\tau = \{ {\rm concat}, {\rm add} \}$ applied at the input level $\ell = \gX^{\rm S}$.
The defining characteristic of PI is its reliance on rich contextual information $c$ containing explicit task instructions, demonstrations, or examples.
For language models, a typical implementation of input manipulation is $I_{\lambda, {\rm concat}, \gX^{\rm S}} = \left[ \lambda_c; \rvx^{\rm T} \right]$, where $\lambda_c$ formats adaptation directives into appropriate instructional text.
For visual tasks, instructions may be incorporated through visual templates or conditional inputs like markers and bounding boxes highlighted on the target input image.
Usually, the output alignment mappings $O_{\omega}$ in PI tasks involve rule-based post-processing such as structured parsing, template filling, or format enforcement.
These operations can be generally expressed as $O_\omega(\rvy^{\rm S}) = \omega(\rvy^{\rm S})$, where $\omega(\cdot)$ denotes extraction or transformation functions that isolate task-relevant information from the model's output.
It can also be identity mapping when no additional output constraints are applied.

By viewing these methods through Eq.~(\ref{eq: reprogrammed_neural_net}), we see they are \emph{not} fundamentally different paradigms but rather specific implementations of reprogrammability, primarily distinguished by their choices for the learnability and location ($\ell$) of $\lambda$, the nature of $c$, and the structure of $\omega$.

\begin{figure}[t]
    \centering
    \includegraphics[width=0.9\linewidth]{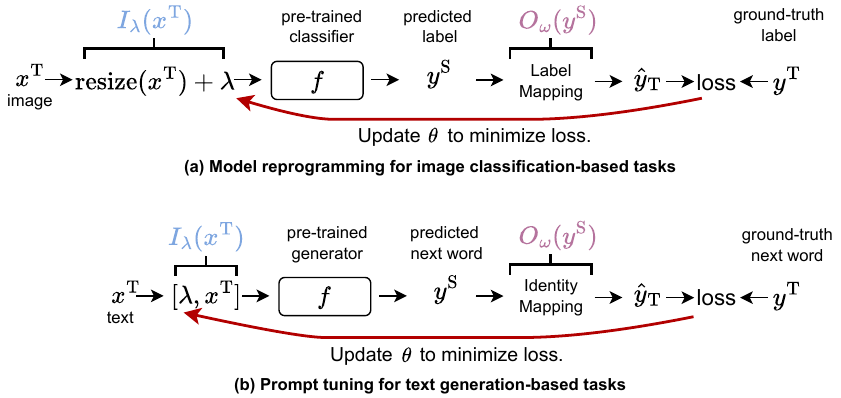}
    \caption{
    Examples of how {\em reprogrammability} manifests across different adaptation methodologies, data modalities, and downstream tasks.
    (a) MR that repurposes a pre-trained image classifier for a new image classification task.
    (b) PT that repurposes a pre-trained language generator for a new text generation task.
    }
    \label{fig:nnr_example}
\end{figure}

\subsection{Reprogrammability as a Unified Paradigm}

The previous sections reveal how MR, PT, and PT share fundamental mechanisms despite their disparate origins.
Building on this, we establish reprogrammability as a broader paradigm that transcends specific techniques, data modalities, and network architectures, bringing conceptual clarity to what have often been disconnected research threads.
Our framework positions MR, PT, and PI as specific instantiations of this general principle.

\paragraph{Shared Principles Across Diverse Adaptation Methods.}
Three fundamental observations highlight the common ground that unites these approaches.
First, the mathematical formulation of Eq.~(\ref{eq: reprogrammed_neural_net}) provides a universal framework regardless of task domain or modality. 
The specific forms of $I_{\lambda, \tau, \ell}$ and $O_{\omega}$ may vary, but their functional roles in transforming inputs and aligning outputs apply universally.

Second, as we will detail in the taxonomy (Sec.~\ref{sec: taxonomy}), the ways $I$ and $O$ can be designed in terms of manipulation formats, locations, operators, and alignment methods form a complete vocabulary for describing \emph{any} method that adapts a model by manipulating its interfaces.
Techniques for different modalities are simply different ``settings'' within this general design space, rather than fundamentally different mechanisms.
For instance, the concept of ``soft prompt'' (continuous learnable tokens) in textual domain has a visual counterpart: one can learn a noise pattern that, when overlaid or concatenated to an image input, guides a pre-trained vision model to a new task (sometimes was termed visual prompt \citep{bahng2022exploring}).

Third, the defining characteristic across all implementations is computational efficiency through interface manipulation rather than parameter modification, which holds regardless of the modalities and architectures involved.
In addition, this principle becomes increasingly valuable as models grow in size. 
The adaptation cost remains relatively constant regardless of model scale, correlating instead with the complexity of the adaptation task itself.

\paragraph{Underlying Mechanisms Enabling Reprogrammability.}

What mechanisms allow a fixed neural network to perform tasks beyond its original training objective? Several complementary perspectives illuminate this phenomenon.

Neural networks' sensitivity to interface perturbations appears to be a universal property and is not tied to a specific architecture.
Both Convolutional Neural Networks~\citep{he2016deep} and transformers~\citep{vaswani2017attention, dosovitskiy2021image}, despite their structural differences, can be repurposed through strategic input manipulations~\citep{chen2023understanding, cai2024sample, cai2024bayesian}.
This sensitivity, related to the high-dimensional, non-linear functions these networks learn (also seen in adversarial examples, Sec.~\ref{sec: adversarial_vulnerability}), means that slight shifts in the input space can traverse significant distances in the representation and output spaces.
Thus, techniques developed for one architecture can often be conceptually adapted to others.

Second, early analyses \citep{yang2021voice2series, chen2024model} suggest that successful reprogramming occurs when sufficient alignment exists between source and target domain representations. 
Input manipulations project target data into regions of the model's representation space where existing computational pathways can be repurposed for new tasks.
Large pre-trained models, in particular, develop high-dimensional vector representations that often exhibit structural similarities and can act as universal function approximators~\citep{lu2020universal}.
Despite processing different modalities, these models develop generalizable computational capabilities that can be activated through appropriate input conditioning. 

Lastly, while a complete theory is still an open research problem, the surprising versatility of large models in diverse tasks through ``prompting'' (e.g., from arithmetic to commonsense reasoning) suggests that they do not just learn one function, but potentially many sub-functions or ``circuits'' relevant to different tasks. 
The RCA approaches, especially prompting, can be seen as ways to activate the specific internal pathways needed for the current task, often guided by mechanisms such as attention, which provide input-dependent activation pathways and dynamically route information based on context~\citep{weiss2022neural}.

\paragraph{Reprogrammability Beyond Text and Vision.} 
The reprogrammability framework extends naturally beyond standard visual and textual domains to emerging modalities. 
As a larger range of sensory modalities has been incorporated into modern AI systems, e.g., audio, tactile~\citep{yen2021neural, hung2023low}, the same principles continue to hold.
\citet{yang2021voice2series} studied how to reprogram acoustic models for time-series data, and \citet{jing2023deep, fang2023universal} adapted pre-trained graph neural networks for molecular property prediction.
In addition, language models are shown to perform protein infilling tasks~\citep{melnyk2023reprogramming,vinod2025reprogramming} and time-series forecasting tasks~\citep{jin2023time}.
These examples illustrate how the underlying formulation of reprogrammability generalizes across fundamentally different data structures and representations.

Moreover, the framework accommodates novel architectures that may emerge in the future. 
As long as a network maintains sensitivity to manipulations at its interfaces--a property that appears intrinsic to neural networks rather than architecture-specific--the reprogrammability principles remain applicable.
This generality suggests that reprogrammability captures fundamental properties of neural network computations.

In particular, the paradigm's applicability becomes valuable in multimodal contexts, where manipulations can target specific modality pathways or cross-modal integration points.
Conceptually, this extends our framework to encompass manipulation locations that span modality boundaries $I_{\lambda, \tau, \ell}^{\text{multi}}(\mathbf{x}^{\rm T}_1, \mathbf{x}^{\rm T}_2, c) = \{I_{\lambda_1, \tau_1, \ell_1}(\mathbf{x}^{\rm T}_1, c), I_{\lambda_2, \tau_2, \ell_2}(\mathbf{x}^{\rm T}_2, c)\}$, where $\rvx_1^{\rm T}$ and $\rvx_2^{\rm T}$ represent different data modalities, such as text and images in vision-language models and multi-modal large-language models.

Thus, we argue that neural network reprogrammability emerges not merely as a collection of domain-specific adaptation techniques but as a paradigm for adapting modern machine learning models.
It provides a unifying perspective on model adaptation that spans modalities, architectures, and application domains, offering both theoretical insights into neural network behavior and practical tools for efficient deployment of pre-trained models across diverse tasks.
\section{A Taxonomy of Reprogrammability}
\label{sec: taxonomy}

\begin{table}[t]
\centering
\caption{
A comparative analysis of existing RCA literature through the lens of reprogrammability.
Under interface ($\ell$), setup $\gX^{\rm S}$, $\gE$, and $\gH$ denote manipulations that occur in input space, embedding space, and hidden space, respectively.
The columns under configuration ($\lambda$): setup OP and FX means optimizable and fixed manipulations.
The columns under operator ($\tau$): setup AD, CO, and PR refer to additive, concatenative, and parametric operators.
The columns under mapping ($\omega$): setup SA, LA, RA, and ID represent statistical alignment, linear alignment, rule-based (i.e., structured) alignment, and identity mapping.
}
\label{tab: framework_manifestation}
\resizebox{0.85\textwidth}{!}{%
\begin{tabular}{@{}lcccccccccccc@{}}
\toprule
\scriptsize
& \multicolumn{8}{c}{\textbf{Input Manipulation ($I$)}} & \multicolumn{4}{c}{\textbf{Output Alignment ($O$)}} \\
\cmidrule(lr){2-9} \cmidrule(lr){10-13}
& \multicolumn{3}{c}{interface ($\ell$)} & \multicolumn{2}{c}{configuration ($\lambda$)} & \multicolumn{3}{c}{operator ($\tau$)} & \multicolumn{4}{c}{mapping ($\omega$)} \\
\cmidrule(lr){2-4} \cmidrule(lr){5-6} \cmidrule(lr){7-9} \cmidrule(lr){10-13}
\textbf{Setup} & $\gX^{\rm S}$ & $\gE$ & $\gH$ & OP & FX & AD & CO & PR & SA & LA & RA & ID \\
\midrule
\textbf{MR} & \checkmark & \checkmark &            & \checkmark &            & \checkmark & \checkmark & \checkmark & \checkmark & \checkmark &            &            \\
\textbf{PT} &            & \checkmark & \checkmark & \checkmark &            &            & \checkmark & \checkmark &            & \checkmark & \checkmark & \checkmark \\
\textbf{PI} & \checkmark &            &            &            & \checkmark & \checkmark & \checkmark & \checkmark &            &            & \checkmark & \checkmark \\
\bottomrule
\end{tabular}%
}
\end{table}

Having established the mathematical framework for neural network reprogrammability, we now present a systematic taxonomy that categorizes adaptation approaches based on their defining characteristics. 
This taxonomy organizes \emph{\textcolor{purple}{reprogrammability-centric adaptation}}~(RCA) approaches along four key dimensions corresponding to design choices in implementing the information manipulation function $I(\cdot; \lambda, \tau, \ell, c)$ and output alignment function $O_\omega$ from Definition~\ref{def: reprogrammability}.
Table~\ref{tab: framework_manifestation} shows a comparative analysis of how these dimensions manifest across existing MR, PT, and PI studies in the literature.
These dimensions are not mutually exclusive but provide a multifaceted structure to understand and compare different approaches.

\subsection{Manipulation Format: Fixed vs. Learnable $\lambda$}

The first dimension distinguishes between RCA approaches based on whether their manipulation configurations $\lambda$ are fixed by manual design or optimizable through learning.
This reflects a core decision between leveraging human expertise versus automated discovery to determine effective manipulation strategies.

\paragraph{Fixed Manipulation: The Power of Human Design.} 
Often, fixed manipulations refer to manually designed, non-learnable manipulations, which leverage human intuition and domain knowledge to craft effective prompts \citep{brown2020language} and instructions \citep{wei2022chain, kojima2022large}.
This approach dominates PI methods, beginning with GPT-3's pioneering demonstrations of in-context learning~\citep{brown2020language}, sparking extensive and rapidly evolving subsequent research into instruction design strategies~\citep{wei2022chain, kojima2022large, mishra2021cross, lu2022learn, zhou2022least}.
Examples include structured reasoning frameworks like chain-of-thought prompting~\citep{wei2022chain}; task decomposition strategies~\citep{kojima2022large}; cross-task instruction transfer~\citep{mishra2021cross} and structured decomposition of reasoning process~\citep{lu2022learn}.

In the visual domain, fixed manipulations take the form of ``hard'' visual prompts, such as points, boxes, or masks. 
Methods like SAM~\citep{kirillov2023segment}, SEEM~\citep{zou2023segment}, and Painter~\citep{wang2023images} illustrate how simple visual cues effectively guide complex visual processing tasks (like segmentation processes) without parameter updates. 
Multimodal systems like CLIP~\citep{radford2021learning} show similar sensitivity to fixed textual prompts--CLIP models could be directed through a textual instruction (e.g., ``a photo of a [{\tt CLASS}]'')--a capability enhanced by techniques like zero-shot prompt engineering~\citep{allingham2023simple}.
Grounding-focused methods like~\citet{koh2023generating} and~\citet{liu2024grounding} leverage text instructions to direct visual attention to specific regions.

Fixed manipulations offer several benefits: they are immediately deployable without training, applicable to black-box models accessible only through APIs, and typically more interpretable than learned alternatives. 
Yet, they typically require more careful domain expertise and/or human intuition, which may not always discover optimal adaptation strategies for complex tasks.

\paragraph{Learnable Manipulation: Optimization-driven Adaptation.}
Learnable manipulation-based RCA approaches, alternatively, treat reprogrammability as an optimization problem, using gradient descent to discover effective manipulation parameters $\lambda$, with a training set $\sD^{\rm T}$ from the target task, such that
\begin{equation}
    \arg \min_{\lambda} \E_{(\rvx^{\rm T}, \rvy^{\rm T}) \sim \sD^{\rm T}} \left[ \gL \left(O_{\omega} I_{\lambda, \tau, \ell} \left( f \left( I_{\lambda, \tau, \ell} (\rvx^{\rm T}, c) \right) \right), \rvy^{\rm T} \right) \right].
\end{equation}\label{eq: learnable_objective}
Here, $\gL(\cdot, \cdot)$ is a task-appropriate loss function (e.g., cross-entropy for classification).

This approach characterizes most MR methods.
Beginning with~\citet{elsayed2019adversarial}, who introduced adversarial reprogramming to repurpose an ImageNet-pretrained image classifier for grid-counting, the field has since expanded to include diverse applications.
Studies have explored cross-domain adaptation~\citep{tsai2020transfer}, continuous vector injection into embedding spaces~\citep{neekhara2019adversarial, hambardzumyan2021warp}, and reprogramming LLMs for protein sequence infilling~\citep{melnyk2023reprogramming}.

PT methods similarly leverage learnable manipulations.
\citet{li2021prefix, lester2021power} showed that optimizing a small set of continuous embedding vectors could match fine-tuning performance in LLMs.
For vision-language models~(VLMs), CoOp~\citep{zhou2022learning} demonstrated that hard-crafted prompts, e.g., ``a photo of a [{\tt CLASS}]'' can be replaced by learnable continuous vectors for more expressive representations.
CoCoOp~\citep{zhou2022conditional} extended this idea to novel classes through image-conditioned prompting.
To further facilitate cross-modal alignment, MaPLe~\citep{khattak2023maple} simultaneously tuned textual and visual modalities, while PLOT~\citep{chen2022plot} minimized distributional discrepancy between modalities that regularizes prompt optimization.

Learnable manipulations typically achieve higher performance than their fixed counterparts, but require access to model gradients, task-specific training data, and computational resources for optimization.

\begin{remark}
    Both fixed and learnable manipulations can be further guided by contextual information $c \in \gC$, which can take various forms.
    For example, we can let $c$ be \emph{sample-specific information} that dynamically adjusts adaptation based on input-specific characteristics~\citep{cai2024sample, qin2021learning, shin-etal-2020-autoprompt} or \emph{domain-specific information}~\citep{gao-etal-2021-making, HAN2022182, Dai2023instruct, cai2025attribute, cai2025understand}.
    In practice, the research trajectory often progresses from fixed manipulations for exploration, progresses to learnable manipulations for performance optimization. 
    The optimal choice depends on a trade-off between performance, resource constraints, and model access.
    Interestingly, with the scaling of model size, fixed manipulations are gaining renewed importance due to the emergent capabilities of large models, as exemplified by in-context learning in LLMs like GPT-4~\citep{achiam2023gpt} and DeepSeek~\citep{liu2024deepseek}, as well as zero-shot vision task capabilities in SAM~\citep{kirillov2023segment}.
\end{remark}

\subsection{Manipulation Location: Where to Modify}
The second dimension categorizes RCA methods based on the location within the model architecture where the manipulation occurs.
This choice influences the level of model access required, the expressivity of the manipulation, and its computational cost.
We identify three primary locations: input, embedding, and hidden spaces.

\paragraph{Raw Input Space $\gX^{\rm S}$ Manipulations.} 
This strategy modifies raw target inputs before they enter the model, preserving the original computational flow while redirecting model behavior toward the target task.
This strategy presents the broadest applicability across model architectures and access scenarios.

Raw input manipulations have been extensively employed in MR~\citep{elsayed2019adversarial, tsai2020transfer, neekhara2019adversarial, englert2022adversarial} for unimodal models and multimodal VLMs~\citep{bahng2022exploring, wu2024evp}.
Similarly, PI methods rely heavily on input space manipulations, as seen in in-context learning demonstrations~\citep{brown2020language}, chain-of-thought~(CoT) reasoning examples~\citep{wei2022chain}, and the multimodal interleaving of visual and textual data in models like Flamingo~\citep{alayrac2022flamingo}.
\citet{tsimpoukelli2021multimodal} demonstrated visual reasoning abilities by providing pre-trained models with appropriate multi-modal examples in the input context.
The adaptation context $c$, in this case, can be exemplified by SMM~\citep{cai2024sample}, which specifies manipulation placements using a sample-wise mask generator.

Since input-level manipulations are universally applicable across diverse models, they are suitable for black-box scenarios where only input-output or API access is available.
They also preserve the original model's computational pipeline, avoiding potential destabilization caused by architectural modifications.
However, directly manipulating discrete input modalities (e.g., text) with continuous data (e.g., learnable tokens) can be difficult.
Additionally, input-level manipulations may struggle to overcome strong biases encoded within the model's internal representations~\citep{zhao2021calibrate}, motivating research into complementary approaches operating at deeper model interfaces.

\paragraph{Embedding Space $\gE$ Manipulations.} 
Embedding space manipulations operate on the model's first learned representation layer, modifying token or feature embeddings before they flow through deeper network layers.
They leverage the model's learned representations while largely preserving the original processing pipeline, but require at least partial white-box access.

These manipulations are frequently used in both MR and PT.
In MR, learned vectors replace certain word embeddings to redirect model behavior~\citep{hambardzumyan2021warp}.
PT approaches like~\citep{lester2021power, li2021prefix} or P-Tuning~\citep{liu2022ptuning} prepended or modified trainable embeddings at strategic positions.
For VLMs, methods like CoOp and CoCoOp learned context textual embeddings that guide visual processing.
CLIP-Adapter~\citep{gao2024clip} incorporated lightweight adaptation layers at the embedding level to improve few-shot performance.
DPT~\citep{zhang2021differentiable} remapped target tasks back to the source domain.
Current trends include improving transferability through methods like multi-task prompt learning~\citep{vu2021spot} and enhancing efficiency with approaches such as mixture-of-prompts~\citep{qin2021learning}.
Mathematically, these manipulations can be expressed as operations on the embedding space $\gE$ of the pre-trained model $f$, either through addition, concatenation, or parametric transformation of embedding vectors.

Embedding-level manipulations offer several advantages: they typically involve few parameters, work particularly well with Transformer architectures~\citep{vaswani2017attention}, and can leverage pre-trained token-level knowledge.
However, they require direct access to the model's internal states, precluding usage in black-box scenarios.

\paragraph{Hidden Representation Space $\gH$ Manipulations.} 
The most invasive form of manipulation targets intermediate activations within the network, requiring full white-box access but potentially offering the most powerful adaptation effects.

While MR rarely operates directly in hidden space, PT methods increasingly leverage these deeper modifications.
P-Tuning v2~\citep{liu2021p} inserted learnable components between transformer layers, showing that deeper placement of prompts leads to more effective reprogramming.
Multi-layer prompting like UniPELT~\citep{he2021towards} and UniAdapter~\citep{lu2023uniadapter} combined manipulations at strategic network depths for enhanced adaptation.
\citet{zhang2024visual} explored the equivalence between visual prompts applied to internal activations and layer normalization.

This approach has proven particularly valuable for cross-modal applications.
For example, VLMs may use task-specific visual adapters that transform features at multiple network depths~\citep{sung2022vl}.
ImageBind~\citep{girdhar2023imagebind} and T-Few~\citep{sanh2022multitask} employed cross-modal alignment layers to transform hidden representations for instruction-based control across multiple input types.

Formally, these manipulations act on the hidden representations $\rvh_i \in \gH_i \subset \gH$ at various depths in the network, modifying the information flow through deeper layers.

\begin{remark}
    The research trend shows increasing interest in targeting deeper model interfaces, particularly for complex cross-domain and cross-modal tasks. However, input-level manipulations remain essential for their universal applicability across diverse model architectures and access scenarios.
\end{remark}

\subsection{Manipulation Operator: How to Transform}
The third dimension examines how RCA approaches integrate manipulations with existing model representations.
This is orthogonal to both location and format, as each operator type can be implemented across different interfaces with either fixed or learnable $\lambda$.

\paragraph{Additive Operators ($\tau = {\rm add}$).}  
These operators superimpose patterns onto existing representations through $I_{\lambda, {\rm add}, \ell} (\rvx^{\rm T}) = \rvx^{\rm T} + \lambda_c$, where $\lambda$ represents manipulation parameters added directly to the input or intermediate representations at location $\ell$.
Additive operators form the basis for many MR approaches \citep{elsayed2019adversarial, bahng2022exploring, neekhara2022cross}.
They are less common in PI, except for visual instructions--bounding boxes~\citep{lin2024draw, chen2023shikra, huang2024a3vlm, ma2024groma, jiang2024joint}, markers~\citep{shtedritski2023does, yang2023set, nasiriany2024pivot}, and pixel-level instructions~\citep{liu2023explicit, zhang2024omg, Amir2022vp}--
include additive visual cues to highlight regions of interest without fundamentally altering the input structure, when combining instructions with input data for new tasks~\citep{kirillov2023segment}.

The mathematical simplicity of additive operators makes them particularly amenable to gradient-based optimization and provides a direct pathway for backpropagation when learning $\lambda$.

\paragraph{Concatenative Operators ($\tau = {\rm concat}$).} 
These operators extend inputs or representations by appending, prepending, or strategically inserting new elements.
In MR, concatenative operators appear in approaches expanding input dimensionality~\citep{wu2024evp, Zhang_2024_CVPR}, such as by padding learnable patterns around target images~\citep{wu2024evp, Zhang_2024_CVPR} or pretending learnable tokens to text input embeddings~\citep{hambardzumyan2021warp}.
Concatenation dominates PT, from regular token prepending \citep{li2021prefix, lester2021power} to dynamic insertion~\citep{qin2021learning}. 
P-Tuning v2~\citep{liu2021p} concatenated prompts at multiple layers. 
Structured Prompting~\citep{hao2022structuredpromptingscalingincontext} incorporated domain knowledge into concatenated prompts.
Concatenative operators also form the foundation of ICL~\citep{brown2020language, wei2022chain} in PI, particularly in CoT prompting~\citep{wei2022chain}, where reasoning steps are appended to demonstrations. 
Moreover, the sensitivity of ICL to the order of concatenated examples has been analyzed~\citep{min2022rethinking, zhang2023what}.

The rise of Transformer-based architectures has particularly amplified the utility of concatenative operators, as these models naturally process sequences of arbitrary length through self-attention mechanisms.

\paragraph{Parametric Operators ($\tau = {\rm param}$).} 
These operators apply complex operations like projections or conditional normalization to inputs or intermediate representations: $I_{\lambda, {\rm param}, \ell} (\rvx^{\rm T}) = \lambda_c(\rvx^{\rm T})$, where $\lambda(\cdot)$ denotes the transformation function.
In MR, parametric operators are used in approaches that apply non-linear transformations of input features for cross-domain adaptation~\citep{tsai2020transfer}.
For PT, parametric operators are strongly associated with adapter-based approaches~\citep{houlsby2019parameter,wang2021kadapt,he2021towards}
Within PI, parametric operators appear in retrieval-augmented~(RAG) approaches~\citep{lewis2020rag,borgeaud2022improving,Trivedi2022InterleavingRW,Dong2023RAFTRR}, which transform inputs by incorporating retrieved documents.

\begin{remark}
    The historical trajectory shows additive operators dominating early work due to their connection to adversarial examples and gradient-based optimization. 
    Concatenative operators gained prominence with the rise of Transformer-based architectures, while parametric operators have recently gained traction as researchers seek more expressive adaptation mechanisms for complex tasks.
\end{remark}

\subsection{Output Alignment Requirements}

The last dimension categorizes RCA approaches based on how they map model outputs to align with target task requirements, reflecting the mechanisms $O_{\omega}: \gY^{\rm S} \to \gY^{\rm T}$ used to interpret pre-trained model's native outputs for the target domain.

\paragraph{Identity Mapping.} 
The simplest form of alignment occurs when model outputs directly correspond to target task outputs without additional processing.
This applies when source and target output spaces are naturally compatible.
Identity mapping is common in PI and PT for unimodal generative tasks, where the model's output text directly serves as the task output. 
As models increase in capability, identity mapping becomes viable for increasingly complex tasks, where visual instruction following through direct generation is demonstrated without post-processing~\citep{liu2023visual, liu2021gpt}. 

\paragraph{Structured Alignment: Rule-based post-processing.} 
Rule-based alignment applies deterministic procedures to extract relevant information from model outputs, enabling structured prediction without additional training. 
Still, it depends on human-designed interpretation rules.
MR rarely employs complex rule-based alignment, since the output spaces between source and target tasks are often significant.
This strategy is central to many PI methods, with CoT~\citep{wei2022chain} and ScratchPad~\citep{Nye2021ShowYW} applying structured parsing to extract reasoning paths from generated text.
ReAct~\citep{yao2023react} and Reflexion~\citep{shinn2023reflexion} interpreted structured outputs like action sequences with pattern-based extraction.
Multimodal instruction models like GPT-4V~\citep{achiam2023gpt} define $O_\omega$ as a pre-defined function that implements parsing, extraction, or validation rules designed to interpret model outputs in task-specific contexts.

\paragraph{Statistical Alignment: Non-parametric Label Mapping.} 
Statistical alignment establishes correspondences between source and target domains based on statistical relationships, without introducing additional trainable parameters.
This approach is predominantly used in the MR literature for classification tasks.
Techniques include random label mapping \citep{elsayed2019adversarial}, frequent label mapping \citep{tsai2020transfer}, iterative maximization of mutual information between source predictions and target categories \citep{chen2023understanding}, and establishing full probabilistic relationships between source and target labels (BLM \citep{cai2024bayesian}), based on the co-occurrence statistics of model predictions (in $\gY^{\rm S}$) and ground-truth labels (in $\gY^{\rm T}$).

\paragraph{Linear Alignment: Parametric Transformation.} 
Linear alignment introduces parametric functions, typically neural networks, to transform source model outputs to target requirements, offering greater flexibility through the optimizable mapping matrix but requiring additional parameters and training data. 
In MR, this corresponds to replacing statistical alignment with a linear probe~\citep{neekhara2022cross}.
In contrast, learnable alignment is more common in PT~\citep{li2021prefix, gu2022ppt}, where a learnable classification head (e.g., verbalizer) was employed atop prompted representations.

\

\begin{remark}
    The choice of alignment strategies reflects the fundamental properties of different adaptation paradigms: statistical and learnable alignments predominate in MR due to the typical mismatch between source and target tasks, while identity and rule-based alignments are more common in PT and PI due to the inherent flexibility of generative models.
    Nowadays, as pre-trained models scale in size and capability, we observe a trend toward simpler alignment strategies, with larger models (especially large-language models) capable of producing outputs that directly address target tasks without complex post-processing, when target tasks also fall under the same modality.
\end{remark}
\section{Emergent Insights from the Reprogrammability Perspective}

Our framework (Sec.~\ref{sec: framework}) and taxonomy (Sec.~\ref{sec: taxonomy}) do more than just catalogue existing methods.
By showing how this perspective illuminates some interesting phenomena, e.g., in-context learning and chain-of-thought reasoning, as well as the trade-offs inherent in different adaptation choices, we can gain insights into how these techniques relate to fundamental properties of the pre-trained model itself.

\subsection{Understanding In-Context Learning (ICL) and Chain-of-Thought (CoT)}
ICL represents one of the most intriguing capabilities of LLMs--the ability to adapt to new tasks through demonstrations without parameter updates--which we can analyze from the viewpoint of our reprogrammability framework, wherein ICL can be characterized as: A non-learnable manipulation ($\lambda \in \Lambda_{\rm fixed}$ determined by demonstration design); Applied exclusively at the input interface ($\ell = \gX^{\rm S}$); Using concatenation ($\tau = {\rm concat}$) to merge demonstrations with queries; and Employing minimal output alignment (typically identity or simple rule-based extraction).
With this formulation, we can reconceptualize ICL as a specific instance of neural network reprogrammability beyond simply a mysterious emergent property, where
\begin{equation*}
    f^{\prime}(\rvx^{\rm T}) = O_{\rm identity} \circ f \circ I_{\lambda_{\rm fixed}, {\rm concat}, \gX^{\rm S}} \left( \rvx^{\rm T}, c_{\rm demonstations} \right).
\end{equation*}
From this perspective, ICL's effectiveness stems from the highly developed interface sensitivity of the pre-trained LLM.
This sensitivity, likely increases along with the scale of pre-training data and model capacity, allows strategic input manipulations (the demonstrations) to dynamically reconfigure or steer the model's internal computational pathways.
A possible reason is the attention mechanisms within Transformer-based architectures, as they allow the model to dynamically weigh the importance of different parts of the input context (including demonstrations) when processing the query. 
The demonstrations $c_{\rm demonstrations}$ ``prime'' the attention patterns and subsequent computations to align with the exemplified task structure.

We argue that this interpretation is well-supported by several findings.
\citet{min2022rethinking} demonstrated that the ICL performance depends critically on aspects like demonstration ordering and input distribution, consistent with the view that concatenative input manipulations create different computational pathways by altering the context seen by the attention layers.
Its effectiveness also scales predictably with model size~\citep{brown2020language}, suggesting that larger models develop more sophisticated interface sensitivity that allows them to better utilize contextual cues.
Recent research suggests that LLMs can learn to implement simple learning algorithms or recognize patterns within their forward pass.
For instance, the emergence of induction heads~\citep{olsson2022context}--attention heads that can complete sequences by copying patterns from the context--provides a concrete mechanism for how models might extract an implicit task-specific transformation function $T_{\rm task}$ from demonstrations $c_{\rm demonstrations}$ and apply it to new inputs $f^{\prime}(\rvx^{\rm T}) \approx T_{\rm task} (\rvx^{\rm T})$~\citep{garg2022can}.
Thus, ICL can be viewed as the model performing a form of rapid, implicit meta-learning or Bayesian inference, where demonstrations act as task-specific ``evidence'' to condition its behavior~\citep{xie2021explanation}.
This way the design of $c_{\rm demonstrations}$ is therefore important, as it directly defines the ``program'' the LLM executes.

CoT reasoning~\citep{wei2022chain} extends this framework by incorporating not just input-output examples, but also structured reasoning steps and intermediate patterns that guide the model through a sequence of computational states.
Within our framework, CoT represents a more specialized and explicit form of input manipulation, where $c_{\rm reasoning}$ now includes explicit reasoning chains:
\begin{equation*}
    f^{\prime}(\mathbf{x}^{\rm T}) = O_{\text{extract}} \circ f \circ I_{\lambda_{{\rm fixed}}, \text{concat}, \mathcal{X}^{\rm S}}(\mathbf{x}^{\rm T}, c_{{\rm reasoning}})
\end{equation*}
where $c_{\rm reasoning} = \left[c_{\rm demonstration}; c_{\rm reason} \right]$ contains demonstration complete with explicit intermediate reasoning steps, and $O_{\rm extract}$ is a potentially more complex output parser that identifies the final answer from the generated reasoning chain.
This ``reprograms'' the model not only to map inputs to outputs, but to follow a particular algorithmic process.
The reasoning steps in $c_{\rm reasoning}$ serve as waypoints or or a scaffold for the model's autoregressive generation process, compelling it to materialize intermediate thoughts that were perhaps only implicitly navigated in standard ICL.


In summary, the reprogrammability perspective shows why CoT can dramatically improve performance on complex reasoning tasks: it explicitly guides the model's computational trajectory through intermediate states that decompose the problem into manageable steps. 
By providing examples of how to reason, CoT leverages the LLM's ability to pattern-match and continue sequences, effectively ``reprogramming'' it to activate or chain together internal circuits responsible for more elementary reasoning capabilities learned during pre-training (e.g., on text containing explanations, arguments, or narratives). 
The very act of verbalizing these intermediate steps, prompted by the CoT examples, can serve as a guidance constraint, ensuring the model follows a more structured and traceable thought process before arriving at an answer. 
This makes the implicit task transformation $T_{\rm task}$ of ICL more explicit and robust, effectively constraining the model to follow sample-specific reasoning patterns.

\subsection{Implications of Taxonomic Position}

Our framework also suggests interesting trade-offs for design choices.

\paragraph{Efficiency-Effectiveness Trade-Offs.}
Different positions in our taxonomy present distinct trade-offs between computational requirements and task performance.
We observe several patterns.
First, input space manipulations generally require fewer computational resources (i.e., $\ll \gO(|\rvx^{\rm T}|)$ parameters) than manipulations performed across multiple hidden layers (i.e., $\gO(\sum^L_{i=1} d_{\rvh_i}$ parameters required).
This difference is amplified in large-scale models with many high-dimensional hidden states.
Second, fixed configurations eliminate the need for training but may sacrifice performance compared to learnable manipulations.
Lastly, the computational cost of different operators varies significantly. Additive manipulations maintain representation dimensionality, concatenative operators increase it (potentially requiring more computation in subsequent layers), and transformative operators add explicit computation steps. 
In practice, the impact of these differences depends on model architecture and the specific implementation.

These trade-offs create distinct adaptation profiles for different resource contexts since learnable RCA methods typically require hundreds to thousands of examples for effective optimization, while well-designed fixed manipulations can adapt with fewer examples. 
When labeled data is scarce, fixed manipulations could be more advantageous despite potentially lower ceiling performance. 
When computational resources are limited but data is abundant, learnable manipulations often provide the best balance of efficiency and effectiveness.

\paragraph{Generalization Properties.}
We also summarize a few findings regarding the generalization of reprogrammed models.
(1) \textit{Sample-specific Adaptation}.
\citet{cai2024sample} observed that manipulations conditioned on input characteristics $I_{\lambda, \tau, \ell}(\rvx^{\rm T}, c(\rvx^{\rm T}))$ tend to generalize more effectively than static manipulations.
This suggests dynamic adaptation better captures the underlying task structure rather than memorizing specific data patterns.
(2) \textit{Representation Hierarchy Effects}.
The manipulation location influences adaptation capabilities.
As layer depth increases, manipulations operate on more abstract representations (i.e., deeper layers) that generalize better across similar tasks but lose modality-specific details. 
This explains why manipulations targeting deeper network layers (particularly within transformer architectures~\citep{liu2021p, he2021towards}) often achieve strong cross-domain transfer.
(3) \textit{Scale-dependent Transfer}.
The generalization gap between input-level reprogramming and fine-tuning follows a power law relationship with model size~\citep{bahng2022exploring}.
That is, input-level reprogramming can underperform fine-tuning with smaller models but achieves competitive results as model size increases
For models below a certain size threshold, fine-tuning consistently outperforms reprogramming. 
Above this threshold, reprogramming approaches competitive or superior performance with significantly fewer parameters~\citep{wu2024evp}, 
e.g., from ResNet~\citep{he2016deep} to CLIP~\citep{radford2021learning}.
Structured task decomposition improves generalization performance on new task-domain combinations~\citep{andreas2022language}.
Recently, \citet{geng2024model} showed that reprogramming is more robust than conventional fine-tuning with out-of-distribution data~\citep{jiang2023detecting, zheng2023out, han2025trustworthy}.

This suggests that adaptation strategies should be tailored to both the available pre-trained model and the expected transfer scenario. 
For large-scale deployment across related tasks, deeper manipulations with sample-specific components offer the strongest performance, whilst for targeted adaptation to a specific task with a smaller model, fine-tuning may remain competitive despite its parameter inefficiency.

\paragraph{Model Scale Dependencies.}
Large language models exhibit emergent reprogrammability in their ability to follow instructions and adapt to new tasks through few-shot demonstrations~\citep{wei2022emergent}, which can be formalized as a phase transition in the model's interface sensitivity:
\begin{equation*}
    \text{Interface Sensitivity}(f, \lambda) \approx \begin{cases}
\epsilon & \text{if } |f| < S_{\text{threshold}} \\
\beta \cdot \log(|f|/S_{\text{threshold}}) & \text{otherwise, }
\end{cases}    
\end{equation*}
where $|M|$ represents model size, $S_{\text{threshold}}$ is the emergence threshold, and $\beta$ is a scaling factor. 
This behavior dictates distinctly different adaptation strategies for models above and below certain scale thresholds, as a sublinear relationship between model size and optimal manipulation volume is observed~\citep{lester2021power}.
Specifically, the proportion of parameters needed for effective adaptation decreases as models grow larger, following an empirically observed relationship, such that $\text{Optimal Prompt Size} \propto \text{Model Size}^{\alpha}$ with $\alpha < 1$.
This sublinear scaling~\citep{lester2021power} suggests that larger models develop more parameter-efficient representational capabilities.

In addition, as model scale increases, the complexity of required output alignment operations decreases~\citep{chowdhery2023palm}. 
For large-scale deployment across related tasks, deeper manipulations with sample-specific components offer the strongest performance, whilst for targeted adaptation to a specific task with a smaller model, fully fine-tuning may remain competitive despite its parameter inefficiency.
\section{Challenges and Future Studies}
\label{sec: challenge_and_future}
Despite the rapid progress and exciting potential, neural network reprogrammability faces open challenges, ranging from theoretical underpinnings to practical deployment.
We outline key issues that future studies should grapple with to unlock the full, responsible potential of this paradigm.

\subsection{Towards a Theory of Reprogrammability}

One of the foremost open questions is \emph{why} and \emph{when} reprogrammability is effective. 
While early theories are emerging~\citep{englert2022adversarial, xie2021explanation, petrov2023prompting, chung2025model}, such as interpreting ICL as a form of implicit Bayesian inference system~\citep{xie2021explanation} or or identifying that pre-trained models learn latent simulators~\citep{hao2023reasoning}, we lack a comprehensive framework predicting a given model’s reprogrammability for a given new task. 
Particularly, what intrinsic properties of a pre-trained model--its architecture, the diversity and scale of its training data, the nature of its learned feature representations (e.g., disentanglement, sparsity), or even its degree of overparameterization—render it amenable to being repurposed for a specific new task via input or context manipulations?
For instance, how do models internally generalize effectively from a few prompt examples, provided that LLMs are known to be able to exhibit ICL abilities?
Current vary, from prompts enabling implicit meta-learning by identifying shared latent structures, to prompts activating specific pre-learned ``skills'' or sub-networks, akin to how attention mechanisms route information.
Is reprogramming essentially unlocking dormant "circuits" within the network, perhaps related to the lottery ticket hypothesis, which posits that sparse, trainable subnetworks exist for many tasks~\citep{frankle2018lottery}? 
Or does high-dimensional overparameterization create such a rich functional landscape that solutions to many tasks are ``accidentally'' embedded and thus discoverable via prompting? 
Crucially, what are the theoretical limits of task complexity that can be handled by reprogrammability without modification to model weights?
Closing this theoretical gap requires delving into concrete questions:
\begin{itemize}
    \item \textbf{Feature Re-utilization vs. New Computation}: To what extent does reprogramming rely on mapping new task inputs to existing feature extractors versus coercing the model into performing novel computations it was not explicitly trained for?
    \item \textbf{Role of Pre-training Objectives}: How do different self-supervised or supervised pre-training objectives (e.g., masked language modeling~\citep{devlin2019bert}, contrastive learning~\citep{radford2021learning}, next-token prediction~\citep{radford2019language}) influence the pre-trained model's reprogrammability for diverse downstream tasks?
    \item \textbf{Theoretical Limits}: What is the upper bound on task complexity or information novelty that a model can adapt to via reprogramming alone, without any weight changes? Can a model be reprogrammed to solve a task whose intrinsic complexity (e.g., related to its Kolmogorov complexity or statistical learning theoretic measures like VC dimension, if the reprogrammed system is viewed as a new learner) exceeds what can be ``described'' by the prompt within the model's existing functional landscape?
    \item \textbf{Predictive Metrics for Reprogrammability}: Can we develop quantifiable metrics, derived from a model's parameters or activation dynamics (e.g., by analyzing its sensitivity to different types of input perturbations), to predict its reprogramming aptitude for a given class of tasks?
\end{itemize}

Answering these questions could profoundly inform how we design and train future foundation models to be inherently more efficient and broadly reprogrammable.

\subsection{Evaluation: Benchmarks and Baselines}

Because reprogrammability enables unconventional use of models, evaluating their performance fairly is tricky. 
If we repurpose an image classifier for a sentiment analysis task via input transformations, what will be an appropriate baseline? 
Is it a model of similar size trained from scratch on the sentiment task, a fine-tuned version of the original image classifier (if feasible), or highly specialized architectures designed for sentiment analysis? 
The choice of baseline dramatically affects perceived efficacy.

A pressing need exists for standardized evaluation frameworks and dedicated benchmarks for RCA methods. 
These are essential for fair comparison and for understanding when RCA should be used rather than parameter-centric adaptations.
Key challenges include:
\begin{itemize}
    \item \textbf{Designing Benchmark Suites}: An urgent need exists for comprehensive benchmark suites specifically for reprogramming. These should encompass a diverse set of source models (varied architectures, sizes, pre-training data); a wide array of target tasks, spanning different modalities (text, vision, multi-modal), varying levels of abstraction, and diverse degrees of similarity/dissimilarity to the source model's original task(s); standardized conditions for evaluation: data limitations (few-shot, zero-shot reprogramming), signal-to-noise ratios, presence of distribution shifts between reprogrammable data and test data, and computational budgets for learning the reprogrammed solution.
    \item \textbf{Defining Appropriate Baselines}: Clear guidelines are necessary for choosing appropriate baselines when evaluating RCA methods. These guidelines should consider factors such as the number of trainable parameters (if any), computational cost (e.g., FLOPs for training and inference), and the type of access to the pre-trained model (e.g., white-box vs. black-box).
    \item \textbf{Setting up Evaluation Metrics}: Evaluations should extend beyond task-specific accuracy. Crucially, reasonable additional dimensions include but are not limited to robustness and stability, program efficiency, computational gains, and transferability of the RCA solution across different tasks.
\end{itemize}
Without rigorous evaluation frameworks, assessing true progress in RCA research will remain challenging, potentially leading to fragmented efforts and incomparable results.


\subsection{Limitations of Existing Programs}

While the concept of reprogrammability is appealing for its efficiency and potential, current techniques, particularly those involving ``prompts'' (whether hard textual prompts or learned soft prompts), still exhibit practical limitations:
\begin{itemize}
    \item \textbf{Data Efficiency}: Although RCA avoids fine-tuning the entire model, learning effective ``soft'' prompts can still demand a considerable amount of target-task data, where the optimization of program parameters can be a challenging non-convex problem, requiring careful initialization and sufficient examples to converge to a good solution.
    Future works may draw inspiration from meta-learning methods~\citep{finn2017model, garnelo2018conditional, garnelo2018neural, ye2022contrastive, du2023idnp, ye2023adversarially, ye2023np, liu2021meta} and domain generalization~\citep{zhong2024domain}, to reduce required sample complexity.
    \item \textbf{Task Complexity and Scope}:
    Another limitation is that current RCA methods excel at tasks that align well with the pre-trained model's inherent capabilities, such as classification, stylistic generation, or question answering, where the underlying knowledge is already present. 
    However, for tasks requiring fundamentally new reasoning chains, complex symbolic manipulation not seen during pre-training (e.g., advanced multi-step mathematics for an LLM not extensively trained on it), or the acquisition of entirely novel skills, merely prepending a prompt or transforming inputs may be insufficient.
    In this case, it is worthwhile to incorporate complex reasoning strategies, e.g., causal reasoning~\citep{chi2024unveiling}, into the reprogramming pipeline.
    \item \textbf{Program Transferability}:
    Moreover, a practical hurdle is that programs, especially learned soft prompts, are often highly specific to the particular model (and sometimes even the specific checkpoint) they were tuned for. 
    A prompt optimized for one LLM architecture rarely transfers effectively to another, even one of similar size or family. 
    When the source model is needed to be adapted for multiple distributionally similar target tasks, this ``hypersensitivity'' restricts the reusability of prompts and forces re-optimization for each new task.
    While research into universal prompts or techniques for efficient prompt adaptation is ongoing, it remains a challenging frontier. 
    Hard-coded prompts, though potentially more transferable, can be brittle and overly sensitive to minor variations in phrasing.
    \item \textbf{Optimization Challenges}:
    MR and soft PT typically rely on gradient access to the model to optimize the prompt parameters. 
    In many real-world scenarios where models are accessed via restricted APIs (black-box access), gradient-based optimization is impossible. 
    This leaves downstream users with manual prompt engineering or less efficient derivative-free optimization methods, such as evolutionary algorithms-based ones~\citep{jiang2016cuckoo}, for discovering effective prompts.
    Developing more powerful and sample-efficient black-box optimization techniques for prompt discovery in high-dimensional spaces is an open problem.
\end{itemize}

Addressing these limitations is key to making RCA a more versatile and robust adaptation solution.

\subsection{Ethical and Security Concerns}
Still, neural network reprogramability introduces a new angle of ethical and security concerns, spanning from misuse to questions of accountability.
As models become more easily redirectable, their potential for unintended or malicious applications grows.

\begin{itemize}
    \item \textbf{Evasion of Safeguards (i.e., prompt injection \& jailbreaking)}: If anyone can repurpose a model via prompts, models could be coaxed into behaviors their creators did not intend~\citep{chen2025refine}. 
    For example, LLMs might be reprogrammed via a cleverly crafted prompt to produce disallowed content, e.g., hate speech, even if safeguards are in place--a form of prompt injection attack~\citep{liu2024autodan, Zou2023UniversalAT}. 
    On the other hand, the reprogrammed models do not possess built-in robustness against malicious manipulations, e.g., adversarial and backdoor attacks~\citep{chen2025refine}. 
    Ensuring that models cannot be easily reprogrammed for harmful purposes via building bespoke defenders, e.g., adversarial detectors~\citep{zhang2025one} and purifiers~\citep{sun2025sample}, is an important yet unresolved challenge.
    \item \textbf{Accountability and Liability}:
    Reprogrammability blurs the lines of responsibility. 
    If a foundation model developed by Company A is reprogrammed by User B using a prompt (perhaps designed or shared by Community C) to generate illegal or harmful content that affects User D, who bears the liability? Is it the original model developer, the prompt designer, or the entity that deployed the reprogrammed system?
    There is an urgent need for clear legal and ethical frameworks for auditing reprogrammed AI systems, attributing actions, and assigning responsibility.
    \item \textbf{Fairness and Bias Propagation}:
    Pre-trained models inevitably inherit biases from their vast training datasets. 
    RCA can inadvertently amplify these biases or introduce new ones when the model is applied to a novel task or data distribution~\citep{Itzhak2024instruct}. 
    For example, a model exhibiting certain gender biases in its original text generation task might manifest these, or even new and unexpected biases, when reprogrammed for a resume screening task in a different demographic context.
    Research is needed into fairness-aware reprogramming techniques, methods to audit reprogrammed models for bias, and understanding how biases transform across tasks and reprogramming methods.
    \item \textbf{Alignment Challenges}: 
    As the reprogrammability increases, developing robust alignment and monitoring techniques becomes essential. 
    There is however a fundamental tension accordingly: excessively rigid constraints or safeguards may stifle legitimate, beneficial adaptability and innovation. Conversely, insufficient guardrails can enable widespread misuse or unintended harmful consequences. The core challenge is achieving alignment not just for the base model, but for the reprogrammed system--ensuring it adheres to the intended goals of the new task while respecting broader safety and ethical principles.
\end{itemize}

Finding the optimal balance between safety and utility represents a central challenge in the responsible deployment of reprogrammable neural networks.
\section{Conclusion}

We presented neural network reprogrammability as a unifying theme connecting model reprogramming, prompt tuning, and prompt instruction--three paradigms enabling us to reuse pre-trained models for new purposes with minimal changes.
We reviewed how this idea developed from early adversarial manipulations into a powerful toolkit for efficient model adaptation, and we offered a unified taxonomy to classify and compare methods across modalities. 
Looking ahead, we anticipate a convergence of ideas: insights from prompting LLMs will inform cross-domain reprogramming in vision and other fields, and vice versa, leading to a more cohesive field of study rather than disparate threads. 
Ultimately, neural reprogramability reflects a shift in mindset: instead of training a new model for each problem, the model itself becomes a platform, and solving a new problem is a matter of programming that platform with the right inputs. 

\bibliography{main}

\clearpage

\end{document}